\pdfoutput=1

\documentclass[11pt]{article}

\usepackage{algorithm}
\usepackage[noend]{algpseudocode}

\usepackage{acl}

\usepackage{times}
\usepackage{latexsym}

\usepackage[T1]{fontenc}

\usepackage[utf8]{inputenc}

\usepackage{microtype}

\usepackage{inconsolata}

\usepackage{graphicx}
\usepackage{booktabs}
\usepackage{amsfonts}
\usepackage{multirow}
\usepackage{enumitem}
\usepackage{multirow}
\usepackage{amsmath}

\title{DAIC-WOZ: On the Validity of Using the \textit{Therapist's prompts} in Automatic Depression Detection from Clinical Interviews}  

\author{Sergio Burdisso\Thanks{\textit{Corresponding authors.} \\
\texttt{\{sergio.burdisso, esau.villatoro\}@idiap.ch}}$^{*,1}$, Ernesto A. Reyes-Ramírez$^2$, Esaú Villatoro-Tello$^{*,1}$,\\ {\bf Fernando Sánchez-Vega}$^{2,3}$, {\bf A. Pastor López-Monroy}$^2$ \and {\bf Petr Motlicek}$^{1,4}$ \\
         $^1$Idiap Research Institute, Martigny, Switzerland \\ $^2$Mathematics Research Center (CIMAT), Gto, Mexico\\
         $^3$Consejo Nacional de Humanidades, Ciencias y Tecnologías (CONAHCYT), México\\
         $^4$Brno University of Technology, Brno, Czech Republic\\ 
         }

\begin{document}
\maketitle
\begin{abstract}
Automatic depression detection from conversational data has gained significant interest in recent years.
The DAIC-WOZ dataset, interviews conducted by a human-controlled virtual agent, has been widely used for this task.
Recent studies have reported enhanced performance when incorporating interviewer's prompts into the model.
In this work, we hypothesize that this improvement might be mainly due to a bias present in these prompts, rather than the proposed architectures and methods.
Through ablation experiments and qualitative analysis, we discover that models using interviewer's prompts learn to focus on a specific region of the interviews, where questions about past experiences with mental health issues are asked, and use them as discriminative shortcuts to detect depressed participants. 
In contrast, models using participant responses gather evidence from across the entire interview.
Finally, to highlight the magnitude of this bias, we achieve a 0.90 F1 score by intentionally exploiting it, the highest result reported to date on this dataset using only textual information.
Our findings underline the need for caution when incorporating interviewers' prompts into models, as they may inadvertently learn to exploit targeted prompts, rather than learning to characterize the language and behavior that are genuinely indicative of the patient's mental health condition.

\end{abstract}

\section{Introduction}
\label{sec:Intro}

Recent advances in Artificial Intelligence (AI) have increased the existing enthusiasm among medical professionals and clinicians when considering the potential for AI-based solutions to make mental healthcare more accessible and to reduce the burden of psychiatric institutions \citep{passos2023digital}. This possibility has led some psychiatrists to argue that the use of AI might result in more standardized and objective measures of mental health \citep{pendse2022treatment}.

Consequently, the automatic analysis of clinical interviews has been recognized as a promising direction for the development of automatic solutions that will help to improve the diagnostic consistency of depression detection \citep{tao23_interspeech,zou2022semi,burdisso2019towards,valstar2016avec}. The Distress Analysis Interview Corpus - Wizard of Oz (DAIC-WOZ) dataset \citep{gratch-etal-2014-distress} stands out as the most representative multimodal resource which has been commonly used for training and validating depression classification models within a clinical setup. Most existing studies leverage the participant answers for depressive assessment, varying from  single-modality methods, i.e., text transcripts, speech \citep{burdisso23_interspeech, villatoroEtAl,Xezonaki2020AffectiveCO, mallolragolta19_interspeech}, to multi-modal approaches (text + speech + video) \citep{chen-naacl-2024, fang2023multimodal,shen2022automatic,yoon2022d,villatorotello21_interspeech}. However, recent studies that incorporate therapist's prompts during training, argue that such information works as supplementary context to better extract salient cues from participant answers \citep{chen-naacl-2024,shen2022automatic,Niu2021, dai2021improving}, reporting high classification performances.

In this paper, we investigate the validity of using the interviewer's prompts from the DAIC-WOZ dataset in automatic depression detection scenarios. We hypothesize that the reported results using both interviewer and participant information may be artificially inflated by a bias induced by the interviewer, failing to generalize to real-world scenarios where such biases may not exist.
The impact of over-reporting performance in the DAIC-WOZ dataset has been already pointed by \citep{bailey2021gender} due to  the presence of gender bias. 
Nevertheless, and to the best of our knowledge, this is the first work to report the existence of a strong bias in the interviewer's prompts and to show that models can effectively exploit it as discriminative shortcuts.

\section{The DAIC-WOZ Dataset}
\label{sec:DAICWOZ_dataset}
The DAIC-WOZ dataset contains clinical interviews in North American English, performed by an animated virtual (human-controlled, i.e., Wizard of OZ) interviewer, called Ellie, designed to support the diagnosis of different psychological distress conditions. The DAIC-WOZ stands as a valuable resource frequently utilized by the NLP community, attributed to its rigorous data collection methods and the scarcity of newer data sources exploring comparable phenomena. 
DAIC-WOZ is a multi-modal corpus, composed by audio and video recordings, and transcribed text from the interviews. To the date, the DAIC-WOZ corpus represents a unique and valuable resource, accumulating over 1K citations since its release.\footnote{Rough estimation based on the citation counts of \citep{gratch-etal-2014-distress, devault2014simsensei} in Google scholar.} 

Ellie conducts semi-structured interviews that are intended to create interactional situations favorable to the assessment of distress indicators 
correlated with depression, anxiety or post-traumatic stress disorder (PTSD). Theoretically, the advantage of Ellie over a human interviewer is the implicit replicability and consistency of the prompts and accompanying gestures. Thus, Ellie has a finite repertoire of 191 prompts, i.e., general questions (\textit{what are you like when you don’t get enough sleep?}), neutral backchannels (\textit{uh huh}),  positive empathy (\textit{that’s great}), negative empathy (\textit{i’m sorry}), surprise responses (\textit{wow!}), continuation prompts (\textit{could you tell me more about that?}), and miscellaneous prompts (\textit{ don’t know; thank you}). Table \ref{tab:dataset} shows a few statistics from the dataset.\footnote{Labels of the test set are not publicly available due to the AVEC competition \cite{valstar2016avec}.}


 \begin{table}[t!]
    \scriptsize
    \centering 
    \begin{tabular}{l@{~~}c@{~~}c@{~~~}c@{~~~}c}
        \toprule
         \textbf{Speaker} & \textbf{Partition} & \textbf{Voc. size} & \textbf{Avg. \#words} & \textbf{Avg. \#tokens}\\
        \midrule
    \multirow{2}{*}{Ellie (\textit{E})} & \textit{train} & 232 & 190.3 ($sd$=26.9) & 567.2 ($sd$=79.10)\\
    & \textit{eval} & 216 & 184.8 ($sd$=50.2) & 540.7 ($sd$=148.5)\\
    \midrule
    \multirow{2}{*}{Participant (\textit{P})} & \textit{train} & 5858 & 621.1 ($sd$=326.2) & 1606.2 ($sd$=893.9)\\
    & \textit{eval} & 3268 & 664.2 ($sd$=281.7) & 1756.3 ($sd$=814.7)\\
        \bottomrule
    \end{tabular}   
    \caption{DAIC-WOZ contains 107 training files (77 control [C] and 30 depressed [D]), an evaluation set of 35 files (23 [C] and 12 [D]). Table shows the vocabulary size and the average interview length measure in words and \textit{WordPiece} tokens, with its corresponding standard deviation (\textit{sd}) values. } 
    \label{tab:dataset}   
\end{table}


\section{Methodology}
\label{sec:method}

To assess the reliability of using Ellie's prompts for automatic depression detection on DAIC-WOZ, we first examine some of the highest results reported in the recent past using this dataset, summarized in Table \ref{tab:results}.
We can categorize published works into two primary groups: (a) those using solely the participant (\textit{P}) responses and, (b) those incorporating Ellie's (\textit{E}) prompts to the model. It seems that works from group (b) exhibit an overall superior performance compared to those of group (a).
To investigate whether this improvement may stem from a bias in Ellie's prompts, before delving into a qualitative analysis, we proposed an initial ablation experiment. 
Concretely, we evaluated two versions of the same models: one employing only participant responses and another solely using Ellie's prompts. Subsequently, we assess the performance difference between these versions, aiming to quantify the challenge in identifying depressed subjects based on participant responses versus Ellie's prompts.
Furthermore, we tested an ensemble approach to measure how complementary these two aspects are to each other.


In particular, we will conduct an ablation experiment using two models: a strong BERT-based baseline model and the Graph Convolutional Network (GCN) model described in \citet{burdisso23_interspeech}, which is the best-performing model that relies solely on the participant's text (see Table\ref{tab:results}).
The choice of these two models aims to compare the baselines against the best-performing model, as well as to analyze models with different natures, namely a bidirectional sequential model and a sequence-agnostic one.
Moreover, as will be described below, the GCN model has an attractive interpretability property that we will use in Section~\ref{sec:analysis} for the qualitative analysis.
Thus, by analyzing the differences between these two models, we can determine whether the observed patterns hold independently of the model's nature. The models are described as follows:

\

\noindent
\textbf{$\bullet$ LongBERT:} a BERT-based classification model. More precisely, we used a pre-trained BERT-based Longformer~\cite{Beltagy2020Longformer} model with a final linear layer added to classify the input using the encoding of the special \emph{[CLS]} token, following common practice.
The choice of using the Longformer variant of BERT~\cite{bert}, instead of the standard Transformer~\cite{transformers} version, stems from the fact that most interviews in DAIC-WOZ are long documents exceeding the 512 token limit (see Table~\ref{tab:dataset}).

\

\noindent
\textbf{$\bullet$ GCN:} The two-layer Graph Convolutional Network (GCN) described in \citet{burdisso23_interspeech} that uses two types of nodes to characterize the interviews: word nodes and participant nodes.
In this graph, nodes are represented at three distinct levels: one-hot encoded vectors, embeddings in a latent space (after applying the first convolution), and in a two-dimensional ``output space,'' (after the second convolution) where each dimension corresponds to the probability of  belonging to the depression or the control group.
Note that since the two type of nodes are represented in the same space, this last learned representation contains probabilities not only for the participants but also for \textit{all the words}.
This is an attractive quality of the model that allows us to track down Ellie's bias to particular subset of words and prompts (as described in  Section~\ref{sec:analysis}).

\section{Experiments and Results}
\label{sec:experimentation}

\begin{table}[t]
    \centering
    \small
    \begin{tabular}{c@{~~~}c@{~~~}c@{~~~}c@{~~~}c@{~~~}c@{~~~}c}
        \toprule
        \multirow{2}{*}{\textbf{Model}} & \multicolumn{3}{c}{\textbf{Source}} & \multicolumn{3}{c}{\textbf{F$_1$ score}}  \\
        & \textit{P} & \textit{E} & \textit{M} & \textit{Avg.} & \textit{D} & \textit{C} \\
        \midrule
        \citet{mallolragolta19_interspeech} & \checkmark &  &  & 0.60 & - & - \\
        \citet{Xezonaki2020AffectiveCO} & \checkmark &  &  & 0.69 & - & - \\
        \citet{villatoroEtAl} & \checkmark &  &  & 0.64 & 0.52 & 0.77 \\
        \citet{burdisso23_interspeech} & \checkmark &  &  & \textbf{0.84} & \textbf{0.80} & \textbf{0.89} \\
        \cmidrule(r){2-7}
        \citet{williamson2016detecting} & \checkmark & \checkmark & & 0.84 & - & - \\
        \citet{Toto2021} & \checkmark & \checkmark & & \textbf{0.86} & - & - \\  
        \citet{shen2022automatic} & \checkmark & \checkmark & & 0.83 & - & - \\
        \citet{Milintsevich2023TowardsAT} & \checkmark & \checkmark &  & 0.80 & - & - \\
        \citet{agarwal2024} & \checkmark & \checkmark &  & 0.77 & - & - \\
        \cmidrule(r){5-7}
        \citet{Niu2021} & \checkmark & \checkmark & \checkmark & 0.92 & - & - \\
        \citet{dai2021improving} & \checkmark & \checkmark & \checkmark & \textbf{0.96} & - & - \\
        \citet{shen2022automatic} & \checkmark & \checkmark & \checkmark & 0.85 & - & - \\
        \citet{chen-naacl-2024} & \checkmark & \checkmark & \checkmark & 0.88 & 0.85 & 0.91 \\
        \midrule
        \midrule
        \textit{P-longBERT} & \checkmark &  &  & 0.72 & 0.64 & 0.80 \\
        \textit{\textbf{E}-longBERT} &  & \checkmark &  & \textbf{0.84} & \textbf{0.80} & \textbf{0.89} \\
        \textit{P-longBERT $\land$ E-longBERT} & \checkmark & \checkmark & & 0.79 & 0.70 & 0.88 \\
        \cmidrule(r){2-7}
        \textit{P}-GCN & \checkmark &  &  & 0.85 & 0.81 & 0.88 \\
        \textit{\textbf{E}}-GCN & & \checkmark &  & \textbf{0.88} & \textbf{0.85} & \textbf{0.91} \\
        \textit{P}-GCN $\land$ \textit{E}-GCN & \checkmark & \checkmark & & \underline{\textbf{0.90}} & \underline{\textbf{0.87}} & \underline{\textbf{0.94}} \\
        \bottomrule
    \end{tabular}
    \caption{Main previously published results on DAIC-WOZ evaluation set along with our obtained results.
    Performance is reported in terms of the F$_1$ score for both control (\textit{C}) and depression (\textit{D}) classes, as well as their macro average (\textit{Avg.}). Results are marked with the source data used: (P) and (E) text from the participant and Ellie; (M) multimodal, e.g., speech and video.
    The global-best result among models using only textual content is \underline{\textbf{underlined}}, while the best results in each group is highlighted in \textbf{bold}.}
    \label{tab:results}
\end{table}

We trained and evaluated two variants of the GCN: one exclusively using the participant's responses as in the original paper~\cite{burdisso23_interspeech}, denoted as \textit{P}-GCN, and another one solely using Ellie's prompts, referred to as \textit{E}-GCN.
Similarly, we also fine-tuned and evaluated the same two versions of the Longformer BERT model, referred to as \textit{P-longBERT} and \textit{E-longBERT}, respectively.\footnote{Details are provided in Appendix \ref{app:implementation}. Source code to replicate our study available at \url{https://github.com/idiap/bias_in_daic-woz}.}
Table \ref{tab:results} shows the obtained results.
When using only the participant responses, \textit{P}-GCN achieved a similarly high F1 score ($0.85$) to the score reported in the original paper ($0.84$), and \textit{P}-longBERT a score ($0.72$) similar to other published works employing solely participant data (e.g. $0.69$). 
On the other hand, when using Ellie, both \textit{E}-GCN and \textit{E-longBERT} achieve comparably higher F1 score.
Notably, \textit{E-longBERT}, by simply utilizing Ellie's prompts, managed to achieve the same score ($0.84$) as the original GCN paper, and the \textit{E}-GCN outperformed all main previously published works that solely rely on textual input, with a score of $0.88$. This suggests that when employing Ellie's prompts, the depression and control groups become more easily distinguishable. For instance, the F1 score of the \textit{longBERTs} for the depression group (D) improves from $0.64$ to $0.80$ when using Ellie's prompts.

Finally, we performed a simple voting ensemble between the two variants of each model, denoted using the ``and'' symbol ($\land$). Participants are classified as positive (\textit{i.e.}, in the depression group) only when both variants, Ellie \textit{and} Participant, classify them as positive.
As shown in Table \ref{tab:results}, the ensemble approach enables the GCN-based model to achieve a remarkable F1 score of $0.90$, the highest reported score to date among models exclusively utilizing textual content.
These results suggest that the integration of both Ellie and participant content could be complementary for certain models, further exploiting Ellie's bias to make the depression and control groups even more easily distinguishable.

\section{Analysis and Discussion}
\label{sec:analysis}

\begin{figure*}[t]
    \centering
    \includegraphics[width=0.8\linewidth]{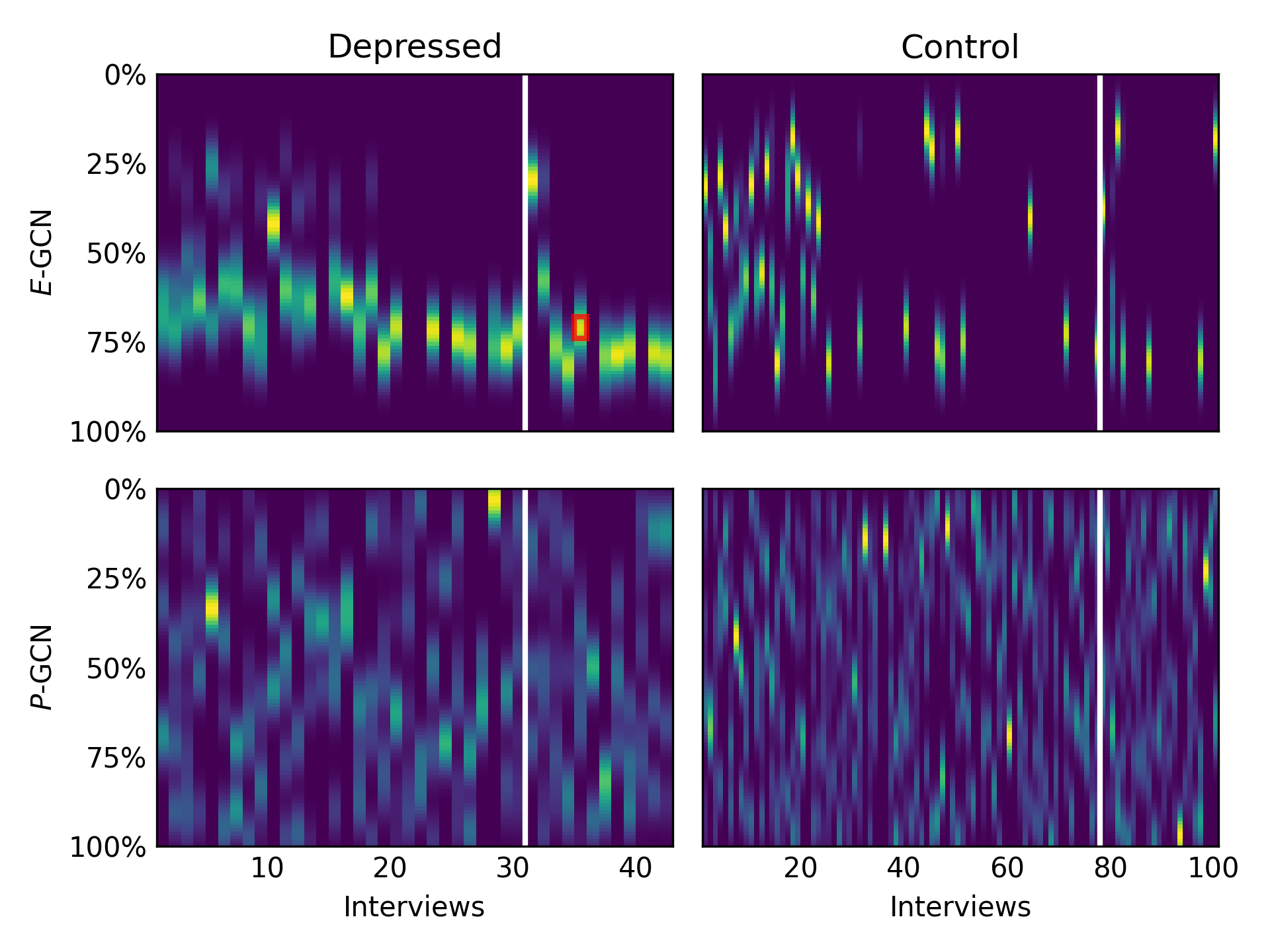}
    \caption{Heatmaps illustrating the distribution of learned keywords by each model across the progression of each interview. The x-axis represents individual interviews, while the y-axis denotes the percentage of the conversation from the beginning (0\%) to the end (100\%). The white vertical line in each plot indicates the training and evaluation splits respectively. Finally, in the \textit{E}-GCN evaluation split region, the small red rectangle depicts the interview segment showed in Fig. \ref{fig:interview-example}.}
    \label{fig:heatmap}
\end{figure*}

Overall, experimental results suggest that Ellie's prompts contain information that the models can exploit to more easily classify the participants.
This is reasonable when considering that therapists adjust their questioning patterns based on the subjects' responses and may adapt their inquiries to delve deeper into specific aspects \textit{when detecting potential depressive symptoms}.

\begin{figure}[t]
    \centering
    \includegraphics[width=\linewidth]{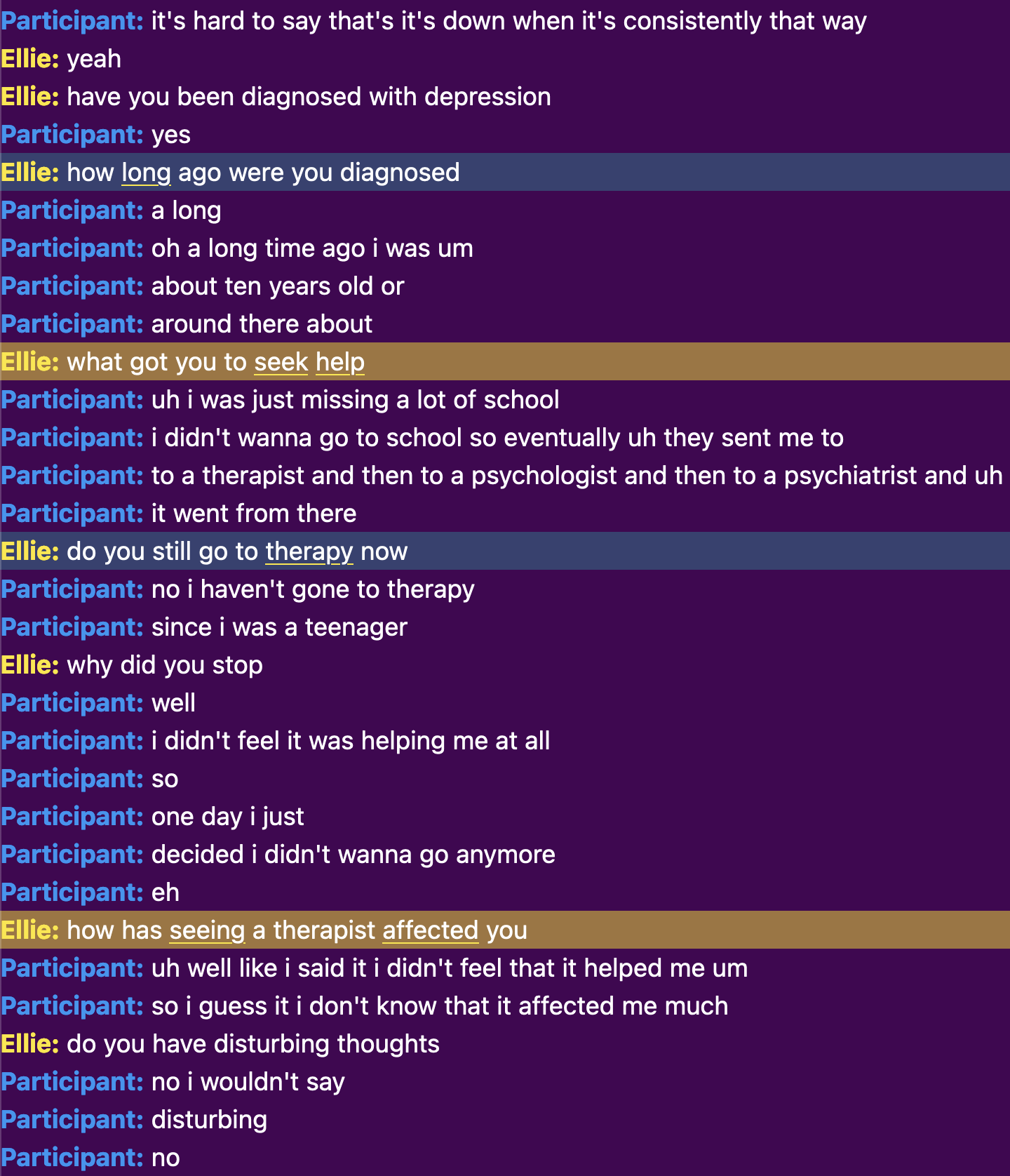}
    \caption{Illustrative segment from interview "381" in the evaluation set, highlighted in Figure \ref{fig:heatmap}. Conversation turns are color-coded based on the proportion of keywords present, with keywords underlined for emphasis.}
    \label{fig:interview-example}
\end{figure}

To explore this possibility further, as mentioned in Section \ref{sec:method}, we leveraged the GCN-based model's ability to learn a common representation for both participant and word nodes in the same output space. Firstly, we extracted the words that both GCN models learned to use to identify the depressed group, which we will refer to as keywords.\footnote{Words \textit{w} such $P(\text{\textit{depressed}}\mid \text{\textit{w}})>P(\neg\text{\textit{depressed}}\mid \text{\textit{w}})$}
Subsequently, we analyzed the distribution of these keywords throughout the progression of each interview to contrast the depressed group against the control group, allowing us to visualize how easily distinguishable the two groups are from the perspectives of both Ellie (\textit{E}-GCN) and the participant (\textit{P}-GCN) models.
Figure~\ref{fig:heatmap} illustrates the distributions obtained from our analysis, highlighting the contrasting behavior of the \textit{E}-GCN and \textit{P}-GCN models. The \textit{P}-GCN distribution exhibits variability across interviews, with no distinct pattern emerging from the distribution of keywords. In contrast, the \textit{E}-GCN model displays a clear and consistent pattern, with concrete regions where keywords concentrate.
That is, the participant model gathers evidence from various parts of the conversations, whereas Ellie's model focuses mainly on very specific segments, \textit{i.e.} specific questions, to classify the participants.
Furthermore, by contrasting the distributions for the depressed group against the control group, we observe that it is easier to distinguish between them using \textit{E}-GCN than \textit{P}-GCN.
This suggests that Ellie's keywords are not only more localized but also possess greater discriminatory power.
Note that for \textit{E}-GCN, in contrast with the control group, almost all the interviews in the depressed group have colored regions, and they are mostly concentrated in a single segment that appears \textit{after halfway the interviews}.\footnote{As shown in Table~\ref{tab:extra-results}, to validate this observation further, we fine-tuned \textit{E-longBERT} on the second half of interviews, achieving $0.84$ F1 (same as full interviews). Using only the first half dropped F1 to $0.60$, highlighting the importance of this latter portion.}
Interestingly, most of these segments correspond to a phase in the interview where Ellie begins to ask more personal questions about past experiences with mental health issues. Figure~\ref{fig:interview-example} shows one such segment. 
Here, we see the segment containing the only four questions that Ellie's model used to classify the participant, disregarding everything else in the conversation, including the question ``\textit{Have you been diagnosed with depression?}'' Note that such questions may be asked to different participants, but an affirmative answer triggered Ellie to delve deeper into specific questions, questions that models could easily learned to identify and exploit to correctly classify the participants.

\subsection{Implications in Clinical Practice}

In clinical practice the final psychiatric diagnosis is typically determined through a clinical interview, often semi-structured, where rating scales serve as additional sources of information to aid in diagnosis. However, these rating scales have limitations, as responses can be influenced by factors such as the patient’s emotional state, comorbidities, relationship with the clinician, and patient self-bias (e.g., participants may be more likely to exaggerate their symptoms \cite{mao2023systematic}). 

Accordingly, the final goal of screening tools such as Ellie, is to contribute towards the replicability, consistency, standardization and the construction of objective measures that support the diagnosis of different mental disorders \cite{pendse2022treatment}.

As shown, the overall analysis described in this paper uncovers interesting biases in the data and shows how ostensibly good performance of NLP models can be deceiving and stress the importance of paying attention to the data and the rationales of the models rather than simply focusing on the superficial performance numbers.
Thus, for automatic depression detection systems to be applicable in real-life clinical practice, systems must be able to provide practitioners whit interpretable and transparent insights to validate systems decisions. There are complex interactions happening during a clinical interview, and accurately modeling is still an open challenge, highlighting the need to develop robust and ethical AI systems for this important and sensitive application domain.   


\section{Conclusions}

Our analysis reveals that the prompts posed by the interviewer, Ellie, contain biases that allow models to more easily distinguish between depressed and control participants in the DAIC-WOZ dataset.
By analyzing the keywords learned by the models, we discover that Ellie's model tends to focus on highly localized segments of the interviews, primarily concentrated in the latter portion where more personal mental health questions are asked. In contrast, the model using participant responses alone does not exhibit such localization, instead gathering evidence from across the entire conversations.
More broadly, our findings underline the need for caution when incorporating interviewers' prompts into mental health diagnostic models. Interviewers often strategically adapt their questioning to probe for potential symptoms. As a result, models may learn to exploit these targeted prompts as discriminative shortcuts, rather than learning to characterize the language and behavior that are truly indicative of mental health conditions.



\section{Ethical Considerations}
In this section, we elaborate on the potential ethical issues. 
\begin{enumerate}
    \item \textbf{Data privacy, participant demographics, and consent.} All the experiments reported in this paper were made on the publicly available DAIC-WOZ dataset, a valuable resource used for training and validating depression detection systems from clinical interviews. This particular dataset was collected by the Institute for Creative Technologies at the University of Southern California. According to the original paper, the DAIC-WOZ dataset received approval from Institutional Ethics Board. All the participants, including the U.S. armed forces veterans and general public from the Greater Los Angeles metropolitan area, were informed that their interviews will be used for academic purposes. All personal details like names, ages, and professions are either removed or anonymized, eliminating any risk of personal information exposure. Original videos from the interviews are not provided, but instead vector features of facial actions and eye gaze are given, making it impossible to reconstruct the participants' appearance. In general, the information of participants was rigorously protected. 
    \item \textbf{The role of AI-based diagnosis.} Our performed experiments aimed at highlighting the importance of using interpretable AI-based solutions as an assistant tools. Thus, the goal is not to replace human experts (psychologists and psychiatrists) but to develop systems that should be used only as support tools. The principle of leaving the decision to the machine would imply major risks for decision making in the health field, a mistake that in high-stakes healthcare settings could prove detrimental or even dangerous. The experiments reported in this paper represent a step forward on the development of bias-aware models in the context of clinical interviews analysis.
\end{enumerate}



\section{Limitations}
 In this section we discuss the limitations of the study described in this paper. 

 \begin{enumerate}
     \item \textbf{Task configuration.} In this paper we only focused on the task of depression detection from clinical interviews, i.e., a controlled scenario where a mental health expert (therapist) conducts an interview with the goal to identify different psychological distress  conditions present in the interviewed participant. This setup is significantly different from the so called ``wild setting'', which refers to the analysis of daily messages, e.g.,  social media posts.  Thus, the findings and claims made in this paper are limited to a clinical setup, and might not be applicable to different setups. As part of our future work, we plan to validate the impact of prompts generated by a fully automatic therapist in similar setups, in particular in the E-DAIC \citep{devault2014simsensei} corpus. 
     \item \textbf{Corpus and modality specific.} Our study is limited to textual modality present in the DAIC-WOZ corpus. Given that the acoustic modality contains also Ellie's interventions, we would like to confirm the presence of the same bias in the acoustic modality. Thus, as part of our future work, we plan to extend our analysis to the additional modalities present in the selected corpus. Similarly, our findings apply specifically to the DAIC-WOZ corpus, hence we cannot confirm the presence of the same type biases in similar corpora. As part of our immediate work, we will replicate our analysis with other datasets like E-DAIC \citep{devault2014simsensei}, EATD \citep{shen2022automatic}, or the recently released ANDROIDS \citep{tao23_interspeech} dataset. 
     
 \end{enumerate}


\section*{Acknowledgements}

This work was supported by Idiap Research Institute's internal funds and computational resources. In addition, we thank CONAHCYT for the computer resources provided through the INAOE Supercomputing Laboratory’s Deep Learning Platform for Language Technologies and CIMAT Bajio Super-computing Laboratory (\#300832). Reyes-Ramírez (CVU 1225869) thanks CONAHCYT for the support through the master's degree scholarship at CIMAT. Sanchez-Vega acknowledges CONAHCYT for its support through the program ``Investigadoras e Investigadores por México'' (Project ID.11989, No.1311).

\bibliography{manuscript}

\newpage
\appendix
\setcounter{table}{0}
\renewcommand{\thetable}{A\arabic{table}}

\section{Technical details}
\label{app:implementation}

\subsection{Graph Convolutional Network}

A Graph Convolutional Network (GCN) is a multilayer neural network that operates directly on a graph and induces embedding vectors of nodes based on the properties of their neighbors. In this work we use the inductive two-layer GCN described in \citet{burdisso23_interspeech}. Let $A\in \mathcal{R}^{n\times n}$ be the weighted adjacency matrix of the graph connecting words and interviews of the DAIC-WOZ training set, the GCN is defined as:

\begin{equation}
    H^{(1)}= \sigma(\tilde{A}H^{(0)}W^{(0)})    
\end{equation}
\begin{equation}
    Z = \text{softmax}(\tilde{A}H^{(1)}W^{(1)})
\end{equation}
where $\tilde{A}=D^{-\frac{1}{2}}AD^{-\frac{1}{2}}$ represents the normalized symmetric adjacency matrix, $W^{(0)}$ is the learned node embeddings lookup table, and $W^{(1)}$ represents the learned weight matrix in the second layer. Loss is computed by means of the cross-entropy between $Z_{i}$ and the one-hot encoded ground truth label $Y_{i}$ for all $i$-th interview in the training set. Following the original paper, we set $k=64$ for the $k$-dimensional feature matrix $H^{(1)}\in \mathcal{R}^{n\times k}$. The adjacency matrix is defined as follows:

\begin{equation}
\label{eq:edge_types}
A_{ij} =
    \begin{cases}
      \text{\textit{mi}}(i,j) & \text{if $i,j$ are words \& \text{\textit{mi}}$(i,j)>0$}\\
      \text{\textit{pr}}(i,j) & \text{if $i,j$ are words \& $i=j$}\\
      \text{tf-idf}_{i,j} & \text{if $i$ is interview \& $j$ is word} \\
      0 & \text{otherwise}
    \end{cases}    
\end{equation}
where \textit{mi} is the point-wise mutual information and \textit{pr} the \emph{PageRank}~\cite{brin1998anatomy} score for node $i$.

Finally, in Section~\ref{sec:analysis} we extracted all the words that the model learned to associate to the depressed category. To select these keywords we selected all words $i$ such that $P(\text{\textit{depressed}}\mid \text{\textit{word}}_i) > P(\text{\textit{control}}\mid \text{\textit{word}}_i)$, that is, $\text{\textit{keywords}} = \{\text{\textit{word}}_i \mid Z_{i,\text{\textit{depressed}}} > 0.5\}$.

\begin{table}[t]
    \centering
    \small
    \begin{tabular}{c@{~~~}c@{~~~}c@{~~~}c|c}
        \toprule
        \textbf{Model} & \textbf{Learning Rate} & \textbf{Epoch} & \textbf{Features} & \textbf{Macro F$_1$} \\
        \midrule
        P-GCN & $1.022\text{e-}06$ & 10 & \textit{top-250} & 0.85 \\
        E-GCN & $1.124\text{e-}06$ & 10 & \textit{auto} & 0.88 \\
        \bottomrule
    \end{tabular}
    \caption{Best hyperparameters obtained for the GCN models after optimization  along with the obtained macro averaged F$_1$ score.}
    \label{tab:extra-results-gcn}
\end{table}

\subsection{Longformer BERT}

The Longformer~\cite{Beltagy2020Longformer} replaces the quadratic self-attention mechanism of Transformers~\cite{transformers} with a combination of global and local windowed attention, scaling linearly with sequence length. This modification enables efficient processing of documents with thousands of tokens, consistently outperforming Transformer-based models on long document tasks.
In particular, we used the version of Longformer described in \citet{chalkidis2022exploration} which has been warm-started re-using the weights of BERT, and continued pre-trained for MLM following the paradigm described in the original Longformer paper. This pre-trained model is available in Hugging Face at \url{https://huggingface.co/kiddothe2b/longformer-mini-1024}.

\begin{table}[t]
    \centering
    \small
    \begin{tabular}{c@{~~~}c@{~~~}c|c}
        \toprule
        \textbf{Model} & \textbf{Learning Rate} & \textbf{Epoch} & \textbf{Macro F$_1$} \\
        \midrule
        P-longBERT & $2.497\text{e-}03$ & 10 & 0.72 \\
        \cmidrule(lr){2-3}
        \textit{first half} & $1.352\text{e-}03$ & \textit{10} & \textit{0.67} \\
        \textit{second half} & $6.051\text{e-}03$ & \textit{10} & \textit{0.73} \\
        \midrule
        E-longBERT & $1.044\text{e-}03$ & 6 & 0.84 \\
        \cmidrule(lr){2-3}
        \textit{first half} & $8.209\text{e-}04$ & \textit{9} & \textit{0.60} \\
        \textit{second half} & $5.075\text{e-}04$ & \textit{7} & \textit{0.84} \\
        \bottomrule
    \end{tabular}
    \caption{Best hyperparameters obtained for the longBERT models after optimization along with the obtained macro averaged F$_1$ score.}
    \label{tab:extra-results}
\end{table}

\subsection{Implementation details}

All models were implemented using PyTorch and were optimized using \emph{Optuna} \cite{akiba2019optuna} with 100 trials for hyperparameter search maximizing the macro averaged F1 score. In each trail, models were trained using AdamW~\cite{loshchilov2018decoupled} optimizer ($\beta_1{=}0.9, \beta_2{=}0.999, \epsilon{=}1\mathrm{e}{-8}$) with \emph{learning rate} and number of epochs $n$ searched in $\gamma \in [1\mathrm{e}{-7}, 1\mathrm{e}{-3}]$ and $n \in [1, 10]$, respectively.
In addition, for GCN, the optimization also tried the three feature selection techniques described in the original paper, \textit{auto}, \textit{top-k}, \textit{none} for, respectively, automatic selection based on term weights learned using Logistic Regression, top-$k$ best selection based on \emph{ANOVA F-value} between words and labels with $k \in \{100, 250, 500, 1000, 1500\}$, and no feature selection (full vocabulary).
Best obtained hyperparameters for the GCN models are shown in Table~\ref{tab:extra-results-gcn}. Finally, Table~\ref{tab:extra-results} presents the parameters obtained for the \textit{longBERT} models, along with the results of the complementary ablation experiments mentioned at the end of Section~\ref{sec:analysis}. Specifically, we divided each interview into two equal parts and performed fine-tuning and evaluation using either the first or the second half. The objective was to reinforce our conclusions regarding the existence of a bias, particularly in the second half of the interviews, as detected by the keywords from the GCN model (Figure~\ref{fig:heatmap}).






\clearpage
\end{document}